\documentclass{article}
\PassOptionsToPackage{numbers}{natbib}

\usepackage[final, nonatbib]{neurips_2025_ml4ps}

\usepackage[utf8]{inputenc} 
\usepackage[T1]{fontenc}    
\usepackage{hyperref}       
\usepackage{url}            
\usepackage{booktabs}       
\usepackage{amsfonts}       
\usepackage{nicefrac}       
\usepackage{microtype}      
\usepackage{xcolor}         
\usepackage[version=4]{mhchem}
\usepackage{multirow}
\usepackage[demo]{graphicx}
\usepackage{pifont}
\usepackage{float}
\usepackage{caption}
\usepackage{subcaption}

\title{Towards Methane Detection On Board Satellites}

\author{Maggie Chen\thanks{These authors contributed equally and are listed in alphabetical order.} \\
  University of Oxford \\
  \texttt{meiqi.chen@physics.ox.ac.uk} \\
  \And
  Hala Lamdouar\footnotemark[1]  \\
  University of Oxford \\
  \texttt{lamdouar@robots.ox.ac.uk} \\
  \And
  Luca Marini\footnotemark[1] \\
  Delft University of Technology \\
  \texttt{l.marini@tudelft.nl} \\
  \AND
  Laura Martínez-Ferrer \\
  Universitat de Val\`encia, Spain\\
  \texttt{laura.martinez-ferrer@uv.es}
  \And
  Chris Bridges \\
  University of Surrey\\
  \texttt{c.p.bridges@surrey.ac.uk}
  \And
  Giacomo Acciarini \\
  European Space Agency (ESA) \\
  \texttt{giacomo.acciarini@esa.int} \\
}

\begin{document}

\maketitle

\begin{abstract}
Methane is a potent greenhouse gas and a major driver of climate change, making its timely detection critical for effective mitigation. Machine learning (ML) deployed on board satellites can enable rapid detection while reducing downlink costs, supporting faster response systems. Conventional methane detection methods often rely on image processing techniques, such as \textit{orthorectification} to correct geometric distortions and \textit{matched filters} to enhance plume signals.
We introduce a novel approach that bypasses these preprocessing steps by using \textit{unorthorectified} data (UnorthoDOS). We find that ML models trained on this dataset achieve performance comparable to those trained on orthorectified data. Moreover, we also train models on an orthorectified dataset, showing that they can outperform the matched filter baseline (mag1c).
We release model checkpoints and two ML-ready datasets comprising orthorectified and unorthorectified hyperspectral images from the Earth Surface Mineral Dust Source Investigation (EMIT) sensor at \url{https://huggingface.co/datasets/SpaceML/UnorthoDOS}, along with code at \url{https://github.com/spaceml-org/plume-hunter}.
\end{abstract}

\section{Introduction}
Methane has a global warming potential approximately 84 times greater than that of carbon dioxide over a 20-year period \cite{myhre2014}. A significant share of point-source methane emissions originates from “super-emitters” in the oil and gas sector, which are detectable from space. With the rising interest in wide-coverage, high-resolution Earth observation (EO), hyperspectral remote sensing imagery provides a unique opportunity to deploy ML models on board spacecraft for rapid greenhouse gas detection. Reliable methane detection at this scale offers a pathway to rapid response and effective mitigation efforts. Methane plume detection in hyperspectral satellite imagery typically involves two key processing steps: (1) \textit{orthorectification}, which corrects geometric distortions caused by sensor viewing angle, terrain variations, and Earth curvature, and (2) the generation of methane enhancement products, often using \textit{matched filters} \cite{manolakis2013detection, thompson2015real, foote2020fast} to strengthen weak plume signals by comparing image data against predefined spectral signatures. Both steps are computationally intensive, making them challenging to execute in resource-constrained environments on board spacecraft. Matched filters are also prone to high false-positive rates. Studies such as \cite{STARCOP1} improved segmentation accuracy by combining matched filter outputs with Red Green Blue (RGB) images in deep learning frameworks. More recently, end-to-end approaches that process hyperspectral bands directly with lightweight neural networks \cite{ruuvzivcka2025hyperspectralvits} further advanced plume detection. As summarized in Table \ref{table:datasets}, these methods remain reliant on orthorectified datasets. However, orthorectification is a costly process to perform in real time on board spacecraft \cite{meoni2024unlocking}. To address this limitation, we present a novel unorthorectified dataset that more closely reflects the data obtained under realistic onboard conditions and demonstrate that methane plumes can also be reliably detected from less-processed satellite imagery.

\begin{table}[tbh!]
  \caption{Comparison of hyperspectral datasets for methane plume detection, with emphasis on the presence or absence of orthorectification preprocessing.}
  \label{table:datasets}
  \centering
  \small
  \begin{tabular}{lcrcc}
  \toprule
  \textbf{Dataset} & \textbf{Instrument} & \textbf{Bands} & \textbf{Spectral range (nm)} & \textbf{\textit{Not} Orthorectified} \\
  \midrule
  STARCOP \cite{STARCOP1} & AVIRIS-NG & 125 & RGB, 1573--1699, 2004--2480 & \ding{55}\\ 
  OxHyperSyntheticCH4 \cite{ruuvzivcka2025hyperspectralvits} & \multirow{3}{*}{EMIT} & \multirow{3}{*}{86} & \multirow{3}{*}{RGB, 1573--1699, 2004--2478} & \ding{55} \\
  OxHyperRealCH4 \cite{ruuvzivcka2025hyperspectralvits} & & & & \ding{55} \\
  UnorthoDOS (Ours) & & & & \ding{51} \\
  \bottomrule
  \end{tabular}
\end{table}

\section{Methodology}


\paragraph{Methane Hyperspectral data}
Our datasets are constructed from hyperspectral observations acquired by the EMIT imaging spectrometer, flown on the International Space Station (ISS) and distributed via the NASA Earthdata portal\footnote{\url{https://search.earthdata.nasa.gov/}}. Specifically, the L1B hyperspectral imagery \cite{L1BEMIT} and L2B methane plume complexes \cite{Methane_plumes}, which serve as ground-truth annotations are used. Both products are provided at a spatial resolution of $60$ m. In total, we curated $1,574$ plume complexes acquired between August 10, 2022, and October 26, 2024. For the purpose of methane detection, a subset of $86$ bands with wavelengths ranging between $1573\text{-}1699$ nm and $2004\text{-}2478$ nm is selected from the original $285$ hyperspectral bands in the L1B images, following the procedure described in \cite{ruuvzivcka2025hyperspectralvits}. This subset captures the methane absorption spectrum, along with non-absorption and RGB bands, enabling ML models to better discriminate methane signals from background features.

\paragraph{UnorthoDOS dataset generation pipeline}

\begin{figure}[tbh!]
    \centering
    \includegraphics[width=\linewidth]{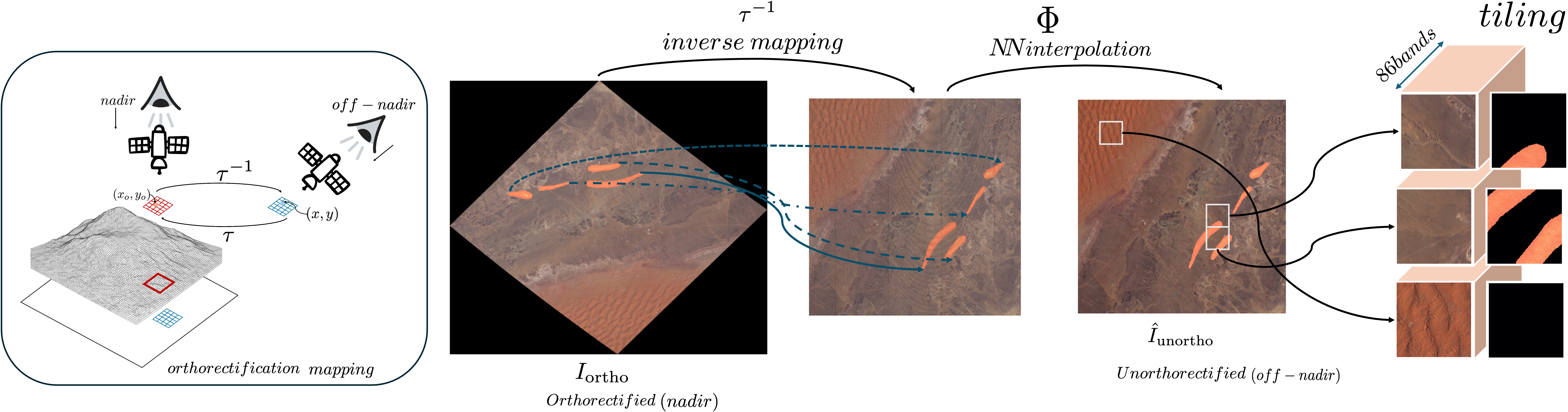}
    \caption{\textbf{Schematic representation of the unorthorectified data generation process}.}
    \label{fig:ortho_transfomation}
\end{figure}

The schematic in Figure~\ref{fig:ortho_transfomation} illustrates the method used for generating our unorthorectified data. 
Orthorectification is defined as a mapping $\tau$ that transforms pixel coordinates $(x, y)$ from an angled (\textit{off-nadir}) image $I_{\text{unortho}}$ to orthogonal (\textit{nadir}) coordinates $(x_o, y_o)$ in the target image $I_{\text{ortho}}$, compensating for sensor geometry and terrain effects (see Figure \ref{fig:ortho_transfomation}). The transformation $\tau$ is extracted from a geometric lookup table in the EMIT satellite data. 
\[
\tau  : \begin{array}{rcl}
I_{\text{unortho}} & \longrightarrow & I_{\text{ortho}}\\
(x,y) & \longmapsto     & \tau(x,y) = (x_o,y_o)
\end{array}
\]
To generate the unorthorectified dataset, the inverse mapping ${\tau}^{-1}$ is approximated by reconstructing $I_{\text{unortho}}$ from $I_{\text{ortho}}$ by using a pixel coordinate grid in the source (unorthorectified) plane, transformed using $\tau$. The transformed grid is then used to sample pixels from the orthorectified plume annotations back in the unorthorectified plane.

For notational simplicity, we denote this operation as ${\tau}^{-1}$, although ${\tau}$ is non-bijective and the images ($I_{\text{unortho}}$ and $I_{\text{ortho}}$) may differ in size. In particular, the inverse mapping ${\tau}^{-1}$ is non-surjective, meaning that certain pixels in $I_{\text{unortho}}$ lack a corresponding value in $I_{\text{ortho}}$; these are filled using nearest-neighbor interpolation. Formally, our procedure for generating approximate \textit{off-nadir} annotations can be expressed as:
\[
\hat{I}_{\text{unortho}} = \Phi \circ {\tau}^{-1} (I_{\text{ortho}}) \text{,}
\]
where $\Phi$ denotes the nearest-neighbor interpolation operator.
Finally, for the purpose of ML training, both the hyperspectral images and their unorthorectified annotations are divided into smaller tiles.

We refer to the resulting unorthorectified dataset as the \textbf{Unorthorectified Dataset for On Board Satellite methane detection (UnorthoDOS)} \cite{UnorthoDOS}.


\paragraph{Orthorectified benchmark}
For comparison, an orthorectified dataset was generated using the orthorectified counterparts of the same L1B images used in the unorthorectified dataset, while retaining the original methane annotations from the L2B products.
The orthorectified images are then added and divided into tiles of $128\times128$ pixels, keeping only those containing over 80\% non-padded pixels.

\begin{figure}[tbh!]
    \centering
    \includegraphics[width=0.9\linewidth,trim=0 48 0 50, clip]{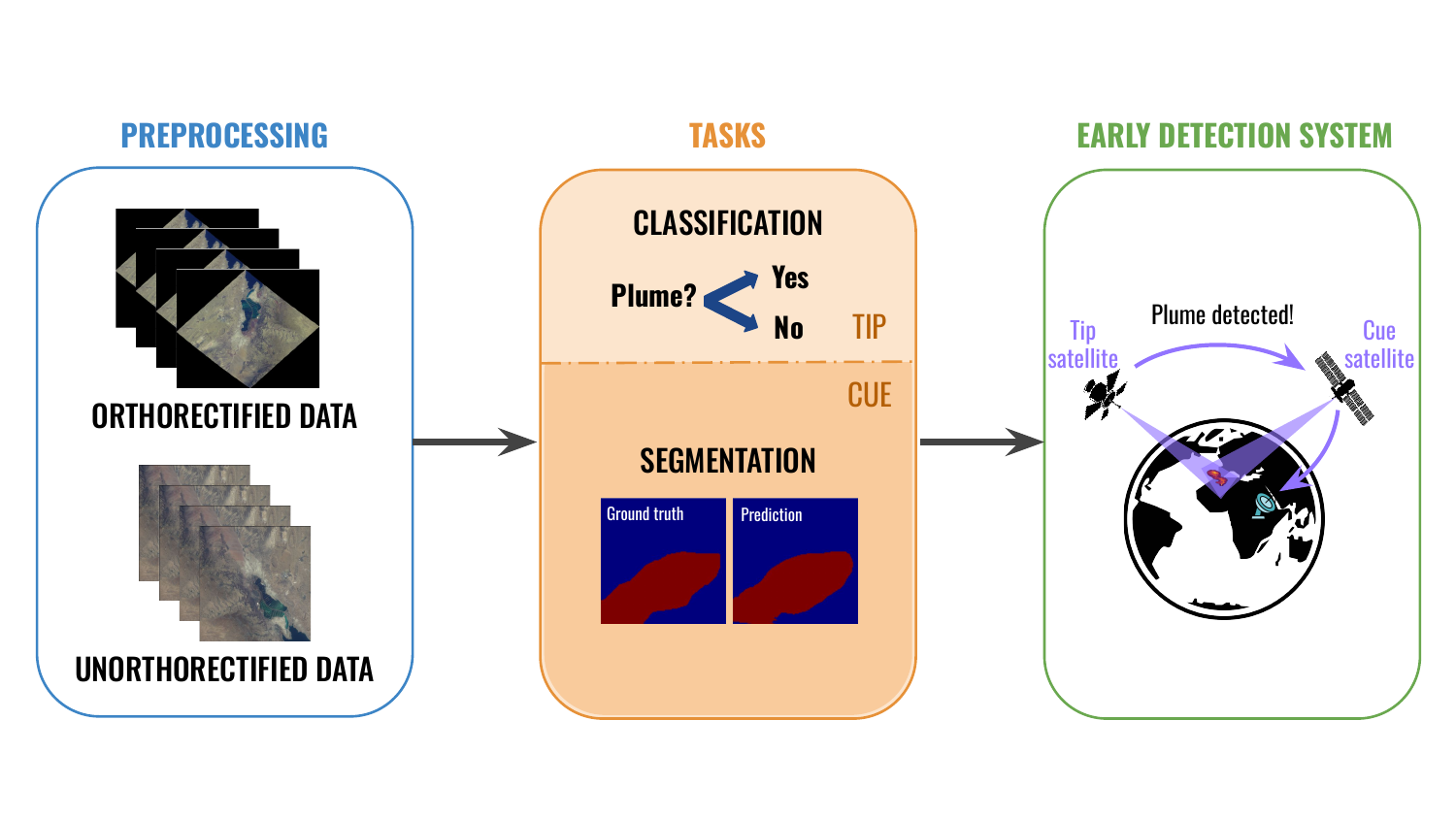}
    \caption{\textbf{Overview of the detection pipeline}. EMIT images are preprocessed into orthorectified and unorthorectified datasets using spatial jittering as data augmentation. 
    The ML models provide an onboard early detection system by performing classification for the TIP satellite and segmentation for the CUE satellite.}
    \label{fig:workflow}
\end{figure}

\paragraph{Experimental setup} 
We present a simplified \textit{tip and cue} paradigm \cite{ceos}, where a tip satellite coarsely localizes methane plumes, and a cue satellite refines the detection.
The tip conducts image classification to determine whether a methane plume is present, while the cue uses semantic segmentation to map its extent (see Figure \ref{fig:workflow}).

The datasets are split to reserve a holdout test set of tiles from 5\% of L1B images, with the rest divided into training (80\%) and validation (15\%) sets. Given the limited number of labelled methane plumes, training and validation sets are augmented via spatial jittering with varying offsets to enhance model generalization. Input tiles are normalized by subtracting the mean and dividing by the standard deviation of each band.

Per-tile methane plume classification and semantic segmentation are performed using a UNet \cite{ronneberger2015u} architecture.
Two models are trained independently on the orthorectified and unorthorectified datasets using a Dice loss, the Adam optimizer \cite{kingma2015adam}, a learning rate of $1\mathrm{e}{-4}$, and a batch size of 32 for 100 epochs. Training is done on a single 40GB NVIDIA A100 GPU.

\paragraph{Evaluation} We report precision, recall, and F1-score for both image classification and semantic segmentation tasks. Additionally, we compute accuracy for image classification, and intersection over union (IoU) for semantic segmentation with model selection based on the highest validation IoU. Using the semantic segmentation model, an image tile is classified as containing a plume if at least one pixel has a positive segmentation prediction.

\section{Results}
Table~\ref{table:results} presents the image classification and semantic segmentation performance of models trained using the orthorectified and unorthorectified datasets, evaluated on image tiles in the corresponding holdout test sets. The performance on strong methane plumes, that is, with a maximum methane concentration threshold of \(\geq 900\) parts per million meter (ppm m), is also evaluated for each model and the mag1c~\cite{foote2020fast} baseline. 

\begin{figure}[tbh!]
\begin{subfigure}{0.49\textwidth}
    \centering
    \parbox[c]{0.32\textwidth}{\centering RGB \& \\ ground truth}%
    \makebox[0.32\textwidth]{UNet}%
    \makebox[0.32\textwidth]{Mag1c}\\
    \begin{minipage}[t]{0.32\textwidth}
        \includegraphics[width=\textwidth]{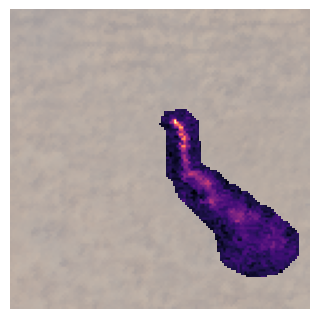}
        \includegraphics[width=\textwidth]{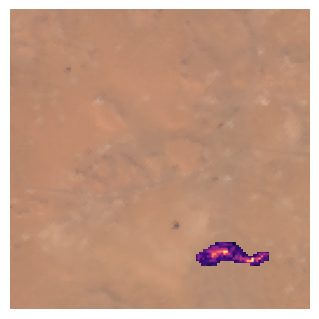}
    \end{minipage}
    \begin{minipage}[t]{0.32\textwidth}
        \includegraphics[width=\textwidth]{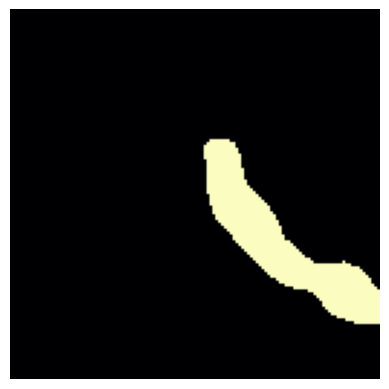}
        \includegraphics[width=\textwidth]{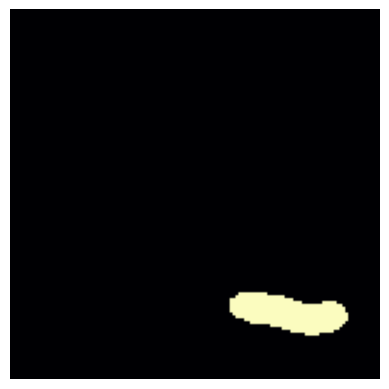}
    \end{minipage}
    \begin{minipage}[t]{0.32\textwidth}
        \includegraphics[width=\textwidth]{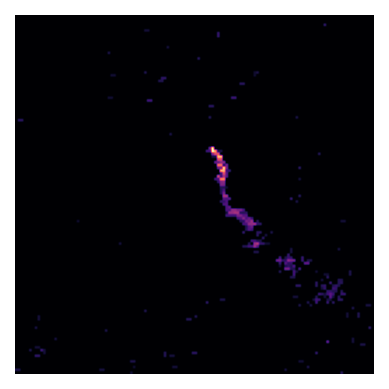}
        \includegraphics[width=\textwidth]{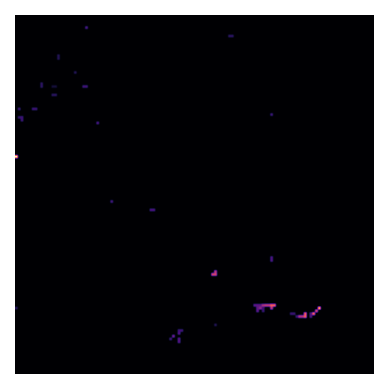}
    \end{minipage}
    \label{fig:qualitative_unortho_fig}
\caption{Orthorectified}
\label{fig:ortho}
\end{subfigure}%
\hspace{0.04\textwidth}
\begin{subfigure}{0.49\textwidth}
    \centering
    \parbox[c]{0.32\textwidth}{\centering RGB \& \\ ground truth}%
    \makebox[0.32\textwidth]{UNet}%
    \makebox[0.32\textwidth]{Mag1c}\\
    \begin{minipage}[t]{0.32\textwidth}
        \includegraphics[width=\textwidth]{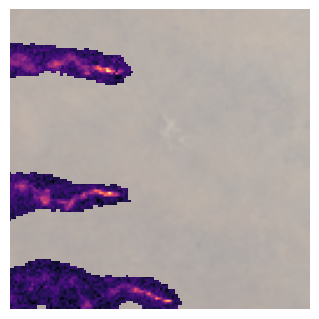}
        \includegraphics[width=\textwidth]{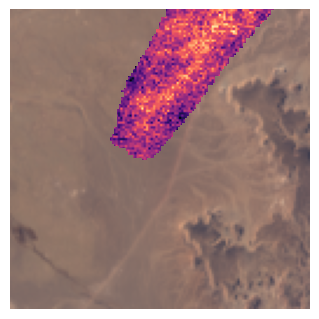}
    \end{minipage}
    \begin{minipage}[t]{0.32\textwidth}
        \includegraphics[width=\textwidth]{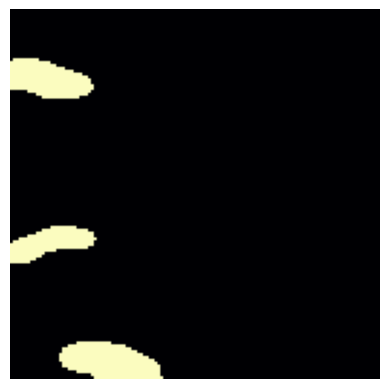}
        \includegraphics[width=\textwidth]{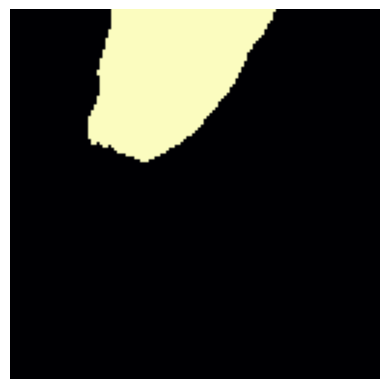}
    \end{minipage}
    \begin{minipage}[t]{0.32\textwidth}
        \includegraphics[width=\textwidth]{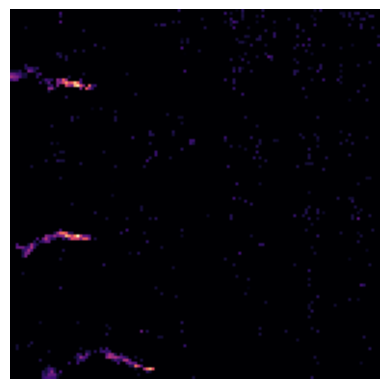}
        \includegraphics[width=\textwidth]{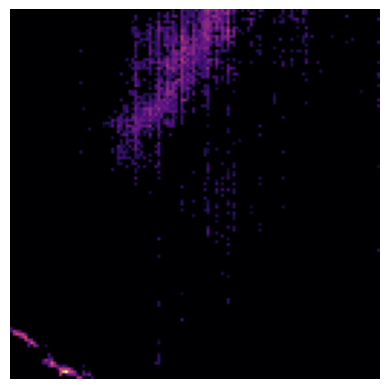}
    \end{minipage}
\caption{Unorthorectified}
\label{fig:unortho}
\end{subfigure}
\caption{\textbf{Visualisation of semantic segmentation results on 2 example tiles from the orthorectified~\ref{fig:ortho} and the unorthorectified~\ref{fig:unortho} datasets each.} From \textit{Left} to \textit{Right} in each sub-figure: L1B tiles (only RGB bands shown for visualisation) overlaid with ground truth methane plume annotations; predicted semantic segmentation plume masks from UNet; segmentation predictions from mag1c~\cite{foote2020fast}}
\label{fig:qualitative_unortho}
\end{figure}

For semantic segmentation, the UNet outperforms the baseline, achieving IoU improvements of 288.03\% for weak plumes and 536.36\% for strong plumes. Figure \ref{fig:qualitative_unortho} shows that UNet more effectively delineates the spatial extent of methane plumes. In contrast, mag1c produces fragmented outputs and exhibits a higher false positive rate, due to the misclassification of terrestrial features as methane emissions. 

Despite UNet’s superior performance over mag1c, absolute performance in the semantic segmentation of weak plumes remains limited, with IoU between 16\% and 19\% in both datasets. Both the segmentation and classification performances improve substantially for strong plumes in tiles with maximum concentration \(\geq 900\) ppm m. For segmentation, the IoU increases by 12.33\% and 9.26\%, respectively, for the orthorectified and the unorthorectified datasets. Whereas for plume classification on strong plumes, the recall increases by 27.12\% for the orthorectified dataset and 22.72\% for the unorthorectified dataset. 

Between the orthorectified and unorthorectified approaches, the performances on both tasks are comparable.
This consistency is observable in Figure \ref{fig:qualitative_unortho}, where the UNet model yields accurate plume predictions for both image types. In the image classification task, the orthorectified dataset yielded slightly superior results, potentially due to differences in spatial alignment of the two datasets or image quality. Conversely, segmentation performance was slightly better for the model trained on the unorthorectified dataset, with the exception of IoU.

\begin{table}[tbh!]
  \caption{Image classification \& semantic segmentation results. Best results per task and dataset are in \textbf{bold}.}
  \label{table:results}
  \centering
  \small
  \begin{tabular}{ccccccccc}
  \toprule
  Task & Model & Threshold (ppm m) & Precision & Recall & F1-Score & Accuracy 
  & IoU \\
  \midrule
  \multicolumn{9}{l}{\textbf{Orthorectified}} \\
  \midrule
  \multirow{5}{*}{Img C} & \multirow{2}{*}{mag1c\cite{foote2020fast}} & N/A & 52.55 & 94.12 & 67.45 & 54.58 & -\\ 
  &  & $\geq$ 900 & 39.60 & \textbf{95.24} & 55.94 & 46.84 & - \\
  & \multirow{2}{*}{UNet} & N/A & \textbf{95.56} & 56.21 & 70.78 & 76.80 &  - \\
  &  & $\geq$ 900 & 94.60 & 83.33 & \textbf{88.61} & \textbf{92.41} & - \\
  \midrule
  \multirow{4}{*}{Sem S} & \multirow{2}{*}{mag1c\cite{foote2020fast}} & N/A & 41.67 & 15.69 & 22.80 & - 
  & 4.76 \\ 
  &  & $\geq$ 900 & 44.91 & 15.69 & 23.25 & 
  - & 4.84 \\
  & \multirow{2}{*}{UNet} & N/A & 79.23 & 19.91 & 31.82 & - & 18.47 \\
  &  & $\geq$ 900 & \textbf{82.77} & \textbf{32.52} & \textbf{46.69} & - & \textbf{30.80} \\
  
  \midrule
  \multicolumn{9}{l}{\textbf{Unorthorectified}} \\
  \midrule
  \multirow{2}{*}{Img C} & \multirow{2}{*}{UNet} & N/A & \textbf{89.19} & 48.89 & 63.16 & 71.48 & - & \\
  & & $\geq$ 900 & 87.88 & \textbf{71.61} & \textbf{78.91} & \textbf{85.65} & - \\
  \midrule
  \multirow{2}{*}{Sem S} & \multirow{2}{*}{UNet} & N/A & 88.41 & 19.85 & 32.42 & - & 16.91 \\
  & & $\geq$ 900 & \textbf{89.08} & \textbf{33.01} & \textbf{48.16} & - & \textbf{26.17} \\
  \bottomrule
  \end{tabular}
\end{table}

\section{Conclusions}

In this work, we present UnorthoDOS, a novel dataset of unorthorectified satellite imagery designed to support ML-based methane plume detection for on board satellite deployment. We demonstrate that models trained on orthorectified and unorthorectified datasets achieve comparable performance, establishing that plume detection is feasible without orthorectification: a critical advantage for real-time detection in resource-constrained satellite scenarios. A key limitation of our study is the the reduced sensitivity to weak methane plumes, which could be substantially mitigated by access to larger annotated hyperspectral image datasets, soon to be enabled by the growing number of satellite missions carrying dedicated sensors.
\section*{Acknowledgments}
This work has been enabled by FDL Earth Systems Lab (https://eslab.ai/) a public / private partnership between the European Space Agency (ESA), Trillium Technologies and the University of Oxford in partnership with Google Cloud, NVIDIA Corporation, Pasteur Labs and SCAN.  
\bibliographystyle{unsrt}
\bibliography{main.bib}

\end{document}